\documentclass[runningheads]{llncs}
\usepackage[T1]{fontenc}
\usepackage{graphicx,verbatim}
\usepackage{booktabs} 
\usepackage{multirow}
\usepackage{bbm} 
\usepackage{amssymb}
\usepackage{wrapfig} 
\usepackage{array}
\usepackage{graphicx}
\usepackage{subcaption}
\usepackage{marvosym}
\usepackage[marginal]{footmisc}

\usepackage{bm}
\usepackage{amsmath}
\usepackage{hyperref}
\usepackage{grffile}
\usepackage{pifont} 
\usepackage{mathtools}
\usepackage{tabularx}
\usepackage{float}
\usepackage{booktabs} % For formal tables
\usepackage{color}
\usepackage{adjustbox}
\usepackage{xcolor}    % 用于设置颜色
\usepackage[table]{xcolor}

\newcommand{\cmark}{\textcolor{green!60!black}{\ding{51}}} 
\newcommand{\xmark}{\textcolor{red!80!black}{\ding{55}}}
\begin{document}
\title{MedPruner: Training-Free Hierarchical Token Pruning for Efficient 3D Medical Image Understanding in Vision-Language Models}
\titlerunning{MedPruner}
% \begin{comment}  %% Removed for anonymized MICCAI submission
\author{Shengyuan Liu\inst{1}\textsuperscript{*} \and
Zanting Ye\inst{2,3}\textsuperscript{*} \and
Yunrui Lin\inst{3}\textsuperscript{*} \and
Chen Hu\inst{2,4} \and
Wanting Geng\inst{5} \and
Xu Han\inst{6} \and
Bulat Ibragimov\inst{7} \and
Yefeng Zheng\inst{2}\textsuperscript{\textdagger} \and
Yixuan Yuan\inst{1}\textsuperscript{\textdagger}}
\authorrunning{S. Liu et al.}
\institute{
\begin{tabular}{@{}c@{\quad}c@{}}
\textsuperscript{1}Chinese University of Hong Kong &
\textsuperscript{2}Westlake University\\
\textsuperscript{3}Southern Medical University &
\textsuperscript{4}Jiangnan University\\
\textsuperscript{5}Dalian University of Technology &
\textsuperscript{6}Shanghai Jiao Tong University\\
\multicolumn{2}{c}{\textsuperscript{7}University of Copenhagen}
\end{tabular}\\
\email{yxyuan@ee.cuhk.edu.hk}\\
\email{zhengyefeng@westlake.edu.cn}\\
\textsuperscript{*} Equal contributions.\\
\textsuperscript{\textdagger} Corresponding Author}
% \author{Anonymized Authors}  %% Added for anonymized MICCAI submission
% \authorrunning{Anonymized Author et al.}
% \institute{Anonymized Affiliations \\
%     \email{email@anonymized.com}}
  
\maketitle              % typeset the header of the contribution
\begin{abstract}
While specialized Medical Vision-Language Models (VLMs) have achieved remarkable success in interpreting 2D and 3D medical modalities, their deployment for 3D volumetric data remains constrained by significant computational inefficiencies. Current architectures typically suffer from massive anatomical redundancy due to the direct concatenation of consecutive 2D slices and lack the flexibility to handle heterogeneous information densities across different slices using fixed pruning ratios. To address these challenges, we propose MedPruner, a training-free and model-agnostic hierarchical token pruning framework specifically designed for efficient 3D medical image understanding. MedPruner introduces a two-stage mechanism: an Inter-slice Anchor-based Filtering module to eliminate slice-level temporal redundancy, followed by a Dynamic Information Nucleus Selection strategy that achieves adaptive token-level compression by quantifying cumulative attention weights. Extensive experiments on three 3D medical benchmarks and across three diverse medical VLMs reveal massive token redundancy in existing architectures. Notably, MedPruner enables models such as MedGemma-1.5 to maintain or even exceed their original performance while retaining fewer than 5\% of visual tokens, thereby reducing visual-token overhead and validating the necessity of dynamic token selection for practical clinical deployment. Our code is available at \href{https://github.com/CUHK-AIM-Group/MedPruner}{here}.

% Authors must provide keywords and are not allowed to remove this Keyword section.

\end{abstract}

\section{Introduction}

%医学多模态模型介绍

Vision-Language Models (VLMs), ranging from proprietary systems like GPT-4o \cite{gpt-4o}, Gemini \cite{gemini} to high-performance open-source models such as the LLaVA \cite{llava} and Qwen-VL \cite{Qwen2.5-VL} series, have demonstrated extraordinary universal perceptual and reasoning capabilities. Drawing upon these advancements, specialized medical VLMs \cite{gigapath,LlavaMed,Lingshu,Huatuo-Vision,medgemma,he2024meddr,llavarad,flamgocxr} such as Med-PaLM M \cite{medpalm} and LLaVA-Med \cite{LlavaMed}, have achieved exceptional proficiency in 2D medical image interpretation, providing critical support for accurate diagnosis and clinical decision-making \cite{endobench,wang2025medical,peng2026omnibrainbench}. Beyond 2D modalities, specialized architectures \cite{m3d,pillar-0,braingpt,med3d-r1,mpLLM} have been developed to navigate 3D volumetric data (e.g., CT and MRI), enabling the analysis of volumetric anatomical structures and temporal dynamics that are critical for complex clinical scenarios. Recently, advanced medical VLMs, such as Hulu \cite{hulumed}, MedGemma-1.5 \cite{medgemma}, and RadFM \cite{RadFM}, have further extended their capabilities to simultaneously process both 2D and 3D medical inputs, aiming to consolidate multimodal medical image understanding within a unified framework. Despite these advancements, extending these models from 2D images to 3D clinical scenarios remains challenging, as high-resolution volumetric medical images can induce a substantial increase in token count. This soaring computational demand necessitates an intelligent token pruning mechanism to maintain reasoning integrity while ensuring clinical-grade inference speeds.

However, current processing pipelines and general-purpose pruning methods exhibit critical limitations when applied to 3D medical inputs. First, their architectures \cite{medgemma,hulumed,Qwen2.5-VL,Lingshu} typically rely on feeding 2D slices along a single axis into the model, where the generated tokens are directly concatenated. Since consecutive slices in 3D volumes share extreme spatial similarity, this direct concatenation introduces massive redundancy that exhausts the LLM’s context window and hinders the processing of auxiliary clinical information. Second, existing token pruning methods \cite{hulumed,visionzip,hiprune,khaki2025sparsevila} typically employ a static, predefined pruning ratio, which fails to account for the inherent heterogeneity of information density. While certain slices capture the intricate boundaries of a tumor, others may only contain uniform tissue with minimal diagnostic value; a fixed ratio either risks losing fine-grained pathological details or wastes tokens on irrelevant backgrounds. Crucially, these static approaches ignore the variance in attention distributions across different vision backbones. Different models exhibit distinct perceptual biases toward medical features, rendering model-agnostic pruning sub-optimal. 

% First, their processing pipelines typically rely on feeding 2D slices along a single axis (e.g., axial, sagittal, or coronal) into the model \cite{medgemma,hulumed,Lingshu,Qwen2.5-VL}. The generated 2D tokens are usually directly concatenated prior to being processed by the Large Language Model (LLM). Since consecutive slices in 3D medical volumes share high anatomical similarity, this direct concatenation introduces massive redundancy that does not contribute to diagnostic decision-making. Such redundancy exhausts the LLM’s context window and limits its capacity to process auxiliary clinical information like patient history. Furthermore, it significantly hinders inference speed, rendering these models impractical for real-time clinical deployment. Second, existing token pruning methods \cite{hulumed,visionzip,hiprune} typically employ a static, predefined pruning ratio, which fails to account for the inherent heterogeneity of information density across different slices. While certain slices capture the intricate boundaries of a tumor, others may only contain uniform tissue with minimal diagnostic value. By overlooking this slice-level heterogeneity, fixed-ratio methods lack the flexibility to adaptively allocate computational resources, leading to sub-optimal efficiency in simple regions or a potential loss of fine-grained pathological information in complex ones.

To address these challenges, we propose MedPruner, a training-free and model-agnostic hierarchical token pruning approach tailored for 3D medical image understanding in VLMs. Specifically, we first incorporate an inter-slice anchor-based filtering module to effectively manage the high temporal redundancy inherent in 3D medical volumes. Additionally, we develop a dynamic information nucleus selection strategy to achieve adaptive compression across slices with varying information densities by quantifying the cumulative attention weights contributed by visual tokens. To validate the efficacy of our proposed method, we conduct extensive experiments on three 3D medical benchmarks and three VLMs. Our results reveal substantial token redundancy in current medical vision-language models. Notably, MedPruner allows MedGemma-1.5 to maintain or even exceed its original performance while using less than 5\% of the original visual tokens. This extreme compression highlights a highly skewed attention distribution in medical VLMs, demonstrating that dynamic token selection is essential for effectively filtering background noise and capturing critical diagnostic signals. Our contributions can be summarized as follows:
\begin{itemize}
\item We introduce MedPruner, which, to the best of our knowledge, is the first training-free and model-agnostic hierarchical token pruning framework tailored to 3D medical VLMs.
\item We employ a training-free, two-stage mechanism to dynamically prune redundant information at both the slice and token levels.
\item We conduct comprehensive experiments across 3 datasets and 3 VLMs, consistently demonstrating the effectiveness and robustness of our approach.
\end{itemize}
% 通过量化 Token 贡献的累积注意力权重，实现对不同信息密度切片的自适应压缩。
\section{Methods}
\subsection{3D Vision-Language Models}
\begin{figure}[t]
  \centering
   \includegraphics[width=\linewidth]{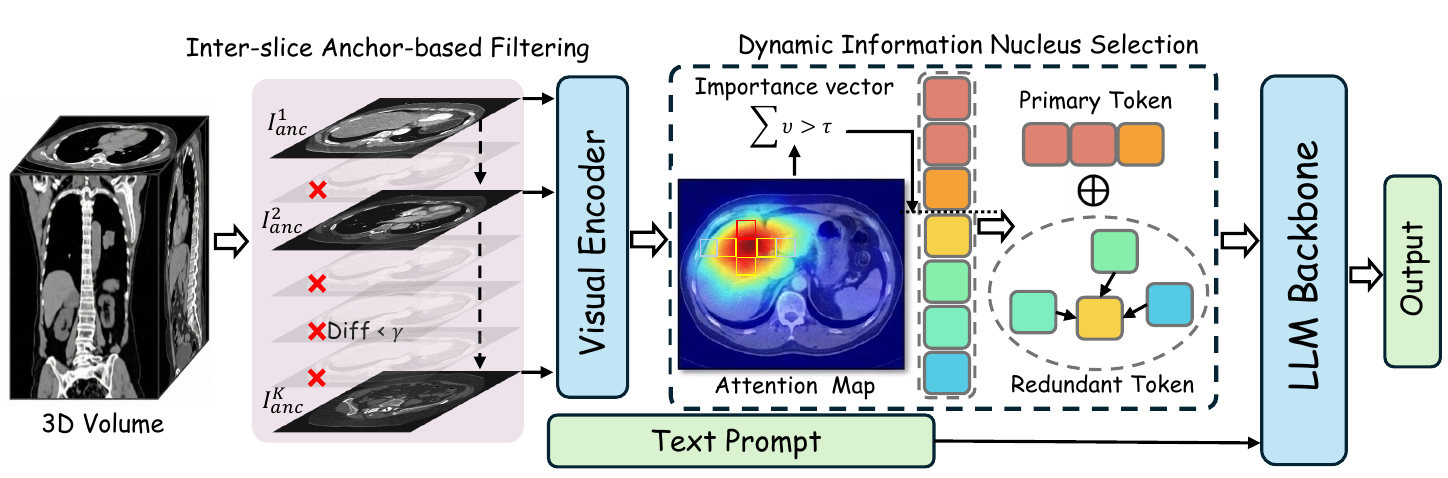}
   \caption{The overview of our MedPruner.}
   \label{fig:main}
\end{figure}

Existing VLM architectures generally consist of three components: a visual encoder, a modality projector, and an LLM backbone. In 3D medical imaging, a common approach is to slice a 3D volume $V \in \mathbb{R}^{D \times H \times W}$ along the axial axis, resulting in a sequence of 2D slices $\mathcal{I} = \{I_1, I_2, \dots, I_D\}$. Each slice $I_i$ is independently processed by a visual encoder $\mathcal{E}_{v}$ and a modality projector $\mathcal{P}$. The final visual representation $H_{v}$ is the concatenation of the projected tokens from all slices:
\begin{equation}
H_{v} = \bigoplus_{i=1}^{D} \mathcal{P}(\mathcal{E}_{v}(I_{i})),
\end{equation}
where $\bigoplus$ denotes sequence concatenation. If each slice yields $M$ tokens, the total visual token count $D \times M$ leads to a severe sequence length explosion. To mitigate this, we propose MedPruner, a hierarchical framework consisting of two components: Inter-slice Anchor-based Filtering (IAF), which reduces inter-slice redundancy by tracking content evolution, and Dynamic Information Nucleus Selection (DINS), which adaptively prunes uninformative tokens based on attention distribution. The overview of MedPruner is shown in Fig.~\ref{fig:main}.

\subsection{Inter-slice Anchor-based Filtering}
To effectively manage the high temporal redundancy inherent in 3D medical volumes, we introduce Inter-slice Anchor-based Filtering (IAF). Rather than employing a static or fixed-interval sampling rate, IAF utilizes a dynamic, content-aware strategy to adaptively identify slices that provide significant anatomical information. The process operates sequentially by maintaining a dynamic anchor slice, denoted as $I_{anc}$. We initialize this filtering process by setting the first slice of the volume as the starting anchor, such that $I_{anc} = I_1$. As we traverse the remaining sequence, we evaluate the informational divergence of each incoming slice $I_i$ ($i > 1$) relative to the current active anchor. This divergence is quantified using the pixel-wise mean $L_1$ distance:
\begin{equation}
\Delta(I_i, I_{anc}) = \frac{1}{N} \sum_{j=1}^{N} |I_{i,j} - I_{anc,j}|,  
\end{equation}
where $N$ denotes the total number of pixels in a slice. This distance $\Delta(I_i, I_{anc})$ serves as a proxy for morphological change. A small distance indicates that the structural information in $I_i$ is already well-represented by $I_{anc}$, rendering the current slice redundant. The core of the IAF mechanism lies in its threshold-driven update logic. We continuously compare the distance $\Delta(I_i, I_{anc})$ against a predefined sensitivity threshold $\gamma$. If the distance exceeds $\gamma$, the slice $I_i$ is deemed to contain significant novel anatomical features. Consequently, $I_i$ is preserved and immediately takes over as the new active anchor ($I_{anc} \leftarrow I_i$) for evaluating the remaining sequence. Conversely, if the distance falls below the threshold, the slice lacks sufficient new information and is entirely filtered out. Through this continuous traversal and updating process, the original dense volume of length $D$ is distilled down to a sparse, informative subsequence comprising only the dynamically preserved anchor frames $\mathcal{I}_{filtered} = \{I_{anc}^1, I_{anc}^2, \dots, I_{anc}^k\}$, $k$ is the final number of retained slices ($k < D$). By discarding the non-anchor intermediate frames, IAF compresses the sequence length. This ensures that the model concentrates its computational budget solely on regions with high structural variance, including the boundaries of organs or the appearance of lesions. Consequently, it prepares a highly condensed and representative sequence for the subsequent token-level optimization.

\subsection{Dynamic Information Nucleus Selection}
Following the inter-slice filtering, we further optimize the token density within each preserved slice by deriving token importance directly from the self-attention layers of the vision encoder. First, we calculate the attention score for each head as follows:
\begin{equation}
S_h = \text{Softmax}\left(\frac{Q_h K_h^\top}{\sqrt{D_h}}\right),
\end{equation}
where $D_h$ is the head dimension, and $Q_h$ and $K_h$ represent the query and key matrices, respectively. By averaging these scores across all heads, we obtain an aggregated attention matrix $S_{\text{avg}}$. To evaluate the raw significance of each visual token, we compute the average of $S_{\text{avg}}$ along the sequence dimension, yielding an initial importance vector $\hat{v} \in \mathbb{R}^{M}$, where $M$ is the number of tokens in the slice. To transform these raw scores into a comparable probability distribution and allow for adjustable selection sensitivity, we apply a temperature-scaled softmax normalization such that:
\begin{equation}
v_i = \frac{\exp(\hat{v}_i / T)}{\sum_{j=1}^{M} \exp(\hat{v}_j / T)},
\end{equation}
where the temperature coefficient $T$ serves as a smoothing factor; a lower $T$ sharpens the distribution to emphasize high-scoring tokens, while a higher $T$ retains broader contextual information. 

The inherent heterogeneity of information density across medical slices suggests that a fixed pruning ratio is suboptimal, as it fails to distinguish between salient anatomical features and uninformative backgrounds. To address this, we utilize a strategy inspired by nucleus filtering to adaptively capture the essential core of each slice by sorting the normalized weights in $v$ in descending order to obtain $v_{\text{sorted}}$. We then dynamically select the minimal set of top-ranked tokens, designated as \textit{primary tokens} $\mathcal{K}$, whose cumulative attention mass reaches a predefined information threshold $\tau$:
\begin{equation}
\mathcal{K} = \{ \text{Top-}k \text{ tokens} \mid \min k \quad \text{s.t.} \quad \sum_{j=1}^{k} v_{\text{sorted}, j} \ge \tau \}.
\end{equation}
By anchoring the selection boundary to the cumulative probability mass, this mechanism ensures that slices with concentrated attention are significantly compressed, while those with dispersed, critical details retain a larger token set to maintain diagnostic integrity. Token dimensions remain unchanged.

Finally, the remaining unselected tokens are treated as \textit{redundant tokens}. To retain the global structural context without increasing the sequence length, we apply a bipartite matching and clustering operation following \cite{visionzip}. These clustered redundant tokens are subsequently concatenated with the primary tokens and passed to the modality projector for the final VLM inference.
\section{Experiments}

\subsection{Experiment Setting}
\noindent \textbf{Datasets.} 
In our experiments, we evaluate our approach on three 3D medical benchmarks: M3D \cite{m3d}, 3D-RAD \cite{3DRad}, and AMOS-MM \cite{Amos}. Both M3D and 3D-RAD are comprehensive 3D Medical Visual Question Answering (VQA) datasets that support both open-ended and closed-ended evaluations. Specifically, M3D queries detailed anatomical and pathological attributes, such as imaging planes, contrast phases, specific organs, and abnormalities. 3D-RAD focuses specifically on radiology CT scans and introduces complex reasoning challenges across six diverse VQA tasks, including anomaly detection, medical computation, and multi-stage temporal diagnosis. Additionally, we utilize the AMOS-MM benchmark, which comprises CT and MRI of abdominal organs, primarily to assess the model's performance in 3D medical report generation.

\noindent \textbf{Implementation Details.} We evaluate MedPruner across three VLMs which support 3D medical imaging input, including a general-purpose model, Qwen3-VL-8B \cite{Qwen2.5-VL}, and two medical domain-specific models, Hulu-Med-7B \cite{hulumed} and MedGemma-1.5-4B \cite{medgemma}. For the evaluation metrics, we utilize Accuracy for closed-set VQA tasks to rigorously measure classification performance. For open-set reasoning and 3D medical report generation, we employ standard Natural Language Generation metrics including BLEU \cite{BLEU} (BLEU-1, BLEU-4), Rouge \cite{Rouge} (Rouge-1, Rouge-L), and METEOR \cite{meteor}. All experiments are conducted on 8 $\times$ NVIDIA H20 GPUs. Hyperparameters are set as $\gamma=0.05$ and $\tau=0.9$.
 
\subsection{Comparison Results}

\begin{table}[t]
\centering
\caption{Quantitative resluts on the 3DRad \cite{3DRad} and M3D \cite{m3d} VQA datasets. \textbf{R-Rate} denotes the token retention rate. \textbf{Bold} and \underline{underlined} values represent the best and second-best performance, respectively.}
\label{tab:m3d_3drad}
\setlength{\tabcolsep}{0.2pt} 
\begin{adjustbox}{width=\textwidth,center}
\begin{tabular}{lcccccccccccc}
\toprule
\multirow{2}{*}{\textbf{Method}} & \multicolumn{6}{c}{\textbf{M3D}} & \multicolumn{6}{c}{\textbf{3DRad}} \\
\cmidrule(lr){2-7} \cmidrule(lr){8-13}
& Acc & Rouge-1 & Rouge-L & BLEU-1 & BLEU-4 & R-Rate & Acc & Rouge-1 & Rouge-L & BLEU-1 & BLEU-4 & R-Rate \\
\midrule
\rowcolor[gray]{0.9} \multicolumn{13}{c}{Hulu-Med-7B \cite{hulumed}} \\
Original                   & 75.317 & 34.188 & 34.047 & 31.820 & 9.541  & 100.00\% & 78.767 & 23.280 & 22.917 & 18.760 & 5.695 & 100.00\% \\
\midrule
Hulu-L1 \cite{hulumed}     & \textbf{77.790} & 47.334 & 47.167 & 45.234 & 12.491 & 66.03\%  & 78.082 & 23.765 & 23.487 & 20.389 & 6.280 & \underline{45.06\%}  \\
VisionZip \cite{visionzip} & \underline{77.623} & \textbf{47.596} & \textbf{47.418} & \textbf{45.451} & 12.549 & 69.96\%  & \underline{79.232} & 25.618 & 25.336 & \underline{22.391} & \underline{6.949} & 69.74\%  \\
HiPrune \cite{hiprune}     & 77.616 & 47.072 & 46.199 & \underline{45.417} & \underline{12.574} & \textbf{22.30\%}  & 79.061 & \underline{25.721} & \underline{25.559} & 21.339 & 6.759 & \textbf{22.30\%}  \\
\textbf{MedPruner}                  & 77.452 & \underline{47.434} & \underline{47.262} & 45.304 & \textbf{12.580} & \underline{52.10\%}  & \textbf{79.280} & \textbf{26.247} & \textbf{25.982} & \textbf{22.865} & \textbf{7.123} & 51.88\%  \\
\rowcolor[gray]{0.9} \multicolumn{13}{c}{MedGemma1.5-4B \cite{medgemma}} \\
Original                   & 32.717 & 6.797  & 5.888  & 3.741  & 0.835  & 100.00\% & 59.155 & 4.865  & 4.094  & 2.629  & 0.516 & 100.00\% \\
\midrule
Hulu-L1 \cite{hulumed}     & \underline{44.638} & 7.810  & \underline{4.473}  & 2.779  & 0.651  & 65.98\%  & \underline{57.868} & 5.009  & 1.596  & 1.034  & 0.217 & 44.89\%  \\
VisionZip \cite{visionzip} & \textbf{45.428} & 7.925  & 4.364  & 2.707  & 0.639  & 69.98\%  & 56.537 & 5.091  & 1.155  & 0.754  & 0.165 & 69.82\%  \\
HiPrune \cite{hiprune}     & 44.420 & \underline{8.201}  & 4.447  & \underline{2.792}  & \underline{0.664}  & \underline{22.30\%}  & 55.959 & \textbf{6.109}  & \underline{2.045}  & \textbf{1.365}  & \textbf{0.304} & \underline{22.30\%}  \\
\textbf{MedPruner}                  & 43.718 & \textbf{8.983}  & \textbf{5.845}  & \textbf{3.923}  & \textbf{1.005}  & \textbf{4.87\%}   & \textbf{60.843} & \underline{5.168}  & \textbf{2.057}  & \underline{1.317}  & \underline{0.277} & \textbf{4.62\%}   \\
\bottomrule
\end{tabular}
\end{adjustbox}
\end{table}

In this section, we conduct a comprehensive evaluation of MedPruner against three existing training-free token reduction methods: the L1-compression method proposed in Hulu-Med \cite{hulumed} (Hulu-L1), VisionZip \cite{visionzip}, and HiPrune \cite{hiprune}. 

Table \ref{tab:m3d_3drad} presents quantitative results on the 3DRad and M3D VQA datasets using Hulu-Med and MedGemma-1.5. In cases where datasets involve massive slice counts, such as M3D with an average of 87 and a maximum of over 600 slices, MedPruner plays a critical role in mitigating information overload. Notably, on the M3D dataset, MedPruner frequently outperforms the uncompressed baseline. This occurs because directly concatenating hundreds of slices introduces significant background noise and redundant structures that can overwhelm the LLM's context window. By filtering these uninformative tokens, MedPruner achieves the highest BLEU-4 scores on M3D (12.580) and 3DRad (7.123) with Hulu-Med while maintaining competitive accuracy and reducing the token retention rate (R-Rate) to approximately 52\%. 

\begin{table}[t]
\centering
\caption{Quantitative results on the AMOS-MM dataset \cite{Amos}. \textbf{R-Rate} denotes the token retention rate. \textbf{Average} represents the mean percentage of performance across all metrics relative to the original model. \textbf{Speed} indicates the average processing time per sample in seconds ($s$). \textbf{Bold} and \underline{underlined} values represent the best and second-best performance, respectively.}
\label{tab:amos}
\setlength{\tabcolsep}{3.5pt} 
\begin{adjustbox}{width=0.9\textwidth,center}
\begin{tabular}{lcccccccc} 
\toprule
Method & Rouge-1 & Rouge-L & BLEU-1 & BLEU-4 & METEOR & Average$\uparrow$ & R-Rate$\downarrow$ & Speed$\downarrow$  \\
\midrule
\rowcolor[gray]{0.9} \multicolumn{9}{c}{Hulu-Med-7B \cite{hulumed}} \\ 
Original                   & 39.518 & 28.763 & 37.317 & 13.375 & 31.729 & 100.00\% & 100.00\% & 9.212 \\
\midrule
Hulu-L1 \cite{hulumed}     & 33.063 & 24.826 & 30.754 & 9.799  & 26.212 & 81.65\%  & \textbf{16.35\%}  & \underline{8.039} \\
VisionZip \cite{visionzip} & 38.519 & \underline{28.216} & \underline{36.876} & \underline{12.705} & 30.738 & \underline{97.25\%}  & 49.69\%  & 8.435 \\
HiPrune \cite{hiprune}     & \underline{39.093} & 27.956 & 35.764 & 12.513 & \underline{31.086} & 96.70\%  & \underline{22.30\%}  & 9.111 \\
\textbf{MedPruner}                  & \textbf{39.255} & \textbf{28.860} & \textbf{37.339} & \textbf{13.120} & \textbf{31.136} & \textbf{99.19\%}  & 54.20\%  & \textbf{7.931} \\
\midrule
\rowcolor[gray]{0.9} \multicolumn{9}{c}{MedGemma1.5-4B \cite{medgemma}} \\ 
Original                   & 13.791 & 9.906  & 5.845  & 0.253  & 9.721  & 100.00\% & 100.00\% & 38.001 \\
\midrule
Hulu-L1 \cite{hulumed}     & \textbf{13.999} & 9.787  & 5.843  & \underline{0.241}  & 9.782  & 99.25\%  & \underline{16.07\%}  & \underline{36.193} \\
VisionZip \cite{visionzip} & \underline{13.830} & \underline{10.043} & \textbf{6.050}  & \textbf{0.260}  & \underline{10.129} & \textbf{102.42\%} & 49.61\%   & 36.682 \\
HiPrune \cite{hiprune}     & 13.369 & 9.782  & 5.828  & 0.236  & 9.939  & 98.21\%  & 21.85\%  & 37.619 \\
\textbf{MedPruner} & 13.706 & \textbf{10.234} & \underline{5.985}  & 0.235  & \textbf{10.252} & \underline{100.65\%} & \textbf{2.46\%}   & \textbf{35.889} \\
\midrule
\rowcolor[gray]{0.9} \multicolumn{9}{c}{Qwen3-VL-8B \cite{Qwen2.5-VL}} \\
Original                   & 18.717 & 18.317 & 26.994 & 1.077  & 18.821 & 100.00\% & 100.00\% & 11.179 \\
\midrule
Hulu-L1 \cite{hulumed}     & 18.201 & 17.808 & 25.839 & 0.870  & 17.788 & 93.10\%  & \textbf{16.50\%}  & \underline{9.569} \\
VisionZip \cite{visionzip} & \underline{18.590} & \textbf{18.047} & 25.604 & \underline{0.925}  & \underline{18.116} & \underline{94.98\%}  & 49.61\%  & 10.654 \\
HiPrune \cite{hiprune}     & 18.217 & 17.255 & \underline{25.886} & 0.919  & \textbf{18.600} & 94.33\%  & \underline{22.30\%}  & 10.052 \\
\textbf{MedPruner}                  & \textbf{19.469} & \underline{17.908} & \textbf{26.119} & \textbf{0.937}  & 17.648 & \textbf{95.86\%}  & 43.28\%  & \textbf{9.044} \\
\bottomrule
\end{tabular}
\end{adjustbox}
\end{table}

Table \ref{tab:amos} reports the performance on the AMOS-MM dataset. Across all tested architectures, MedPruner achieves the optimal balance between accuracy and efficiency, delivering the fastest inference speeds while maintaining exceptional performance. In several instances, it even surpasses the original baselines, such as reaching a 100.65\% average score on MedGemma-1.5. These results highlight the model-agnostic robustness of MedPruner in optimizing diverse VLMs under real-world computational constraints.

A notable observation is the extreme token compression achieved by MedPruner on the MedGemma-1.5 model. As shown in the tables, MedPruner maintains high performance while requiring fewer than 5\% of the visual tokens across all three datasets, reaching an incredibly low R-Rate of 2.46\% on AMOS-MM. Upon analyzing the model's behavior, we discovered that MedGemma-1.5's attention weights are highly concentrated on a single or a very small subset of tokens. By dynamically anchoring the selection threshold to the cumulative attention mass, our module naturally adapts to this highly skewed distribution. This phenomenon strongly validates the necessity of our dynamic selection strategy. Unlike fixed-ratio methods that arbitrarily discard useful tokens or retain unnecessary ones (e.g., HiPrune fixed at 22.30\%), MedPruner intelligently scales the token retention rate based on the intrinsic attention distribution, ensuring optimal efficiency tailored to each specific slice and model architecture.

\subsection{Ablation Study}
% We conduct ablation studies to investigate the contribution of individual components within MedPruner and the sensitivity of the key hyper-parameter $\tau$ to the overall performance.
\begin{figure}[htbp]
    \centering
    % ========== 左侧：表格 ==========
    \begin{minipage}[c]{0.52\textwidth}
        \centering
        \captionof{table}{Ablation study on the AMOS-MM dataset \cite{Amos} using the Hulu-Med-7B model \cite{hulumed}.}
        \label{tab:ablation}
        \setlength{\tabcolsep}{3pt}
        \begin{adjustbox}{width=\linewidth,center}
        \begin{tabular}{cccccc}
        \toprule
        IAF & Primary & Redundant & Average$\uparrow$ & R-Rate$\downarrow$ & Speed$\downarrow$  \\
        \midrule
        \xmark & \xmark & \xmark &  100.00\% & 100.00\%  & 9.212 \\
        \cmark & \xmark & \xmark & 92.13\%  & \underline{60.33\%} & \textbf{7.751}  \\
        \xmark & \cmark & \xmark & 98.73\% & 83.11\% & 8.894\\
        \xmark & \cmark & \cmark & \textbf{100.07\%} & 88.14\% & 9.586 \\
        \midrule
        \cmark & \cmark & \cmark  & \underline{99.19\%} & \textbf{54.20\%} & \underline{7.931}  \\
        \bottomrule
        \end{tabular}
        \end{adjustbox}
    \end{minipage}
    \hfill 
    % ===================
    \begin{minipage}[c]{0.43\textwidth}
        \centering
        \includegraphics[width=\linewidth]{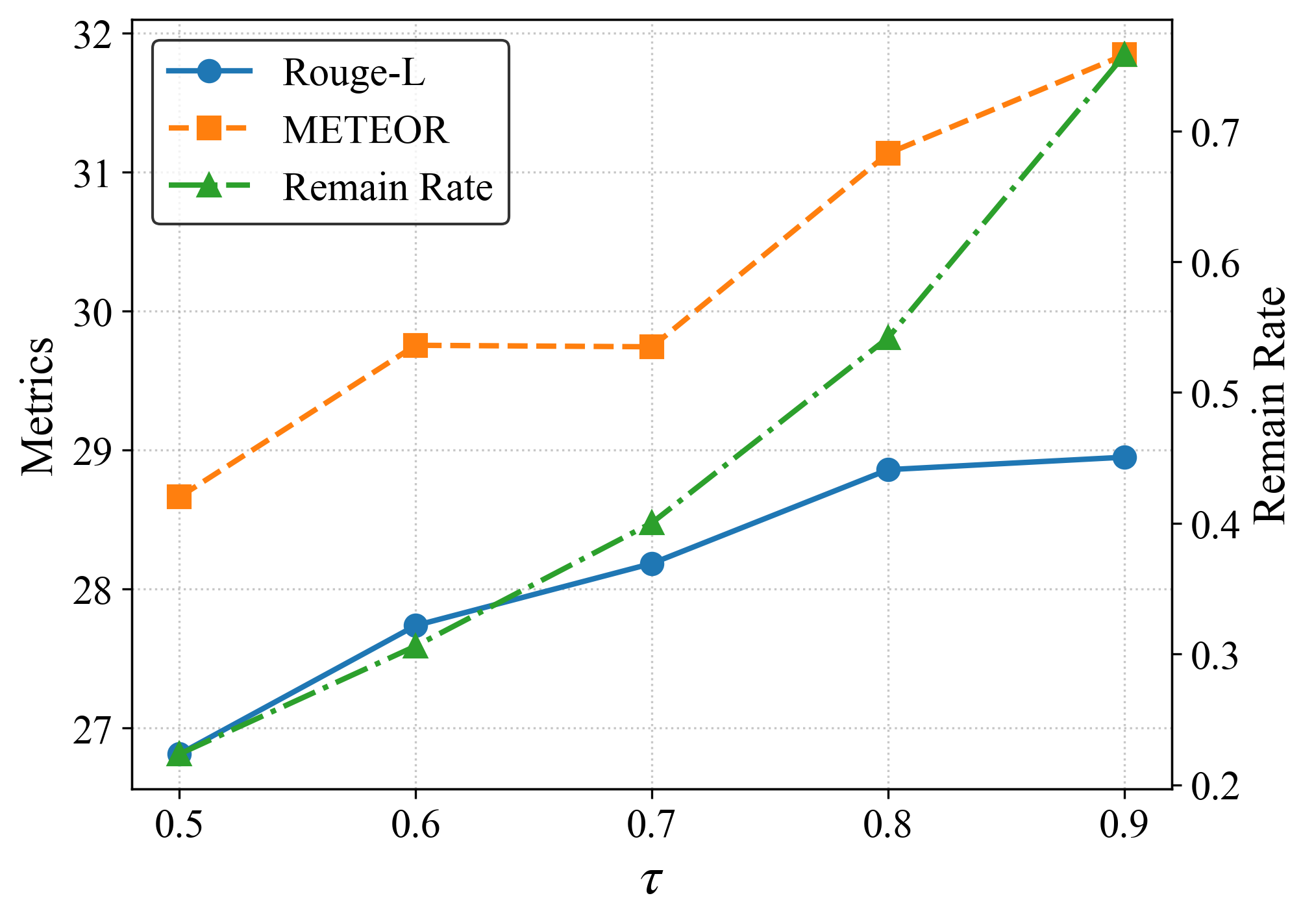}
        \captionof{figure}{Ablation study of $\tau$.}
        \label{fig:tau}
    \end{minipage}
\end{figure}

\noindent \textbf{Component Analysis.} We first evaluate the contribution of each MedPruner component on the AMOS-MM dataset, with results summarized in Table~\ref{tab:ablation}. As shown in Table~\ref{tab:ablation}, IAF lowers average inference time from 9.2s to 7.7s. While this slice-level filtering initially leads to a performance drop, the introduction of Primary token selection and Redundant token clustering effectively restores diagnostic accuracy. Specifically, incorporating primary tokens and redundant clustering recovers the Average score to over 100\% of the baseline. The full MedPruner configuration achieves an optimal balance, maintaining 99.19\% of the original performance while reaching the strongest compression with a 54.20\% R-Rate.

\noindent \textbf{Sensitivity of Information Threshold $\tau$.} Fig.~\ref{fig:tau} illustrates the impact of the information threshold $\tau$ on the Hulu-Med-7B model. As $\tau$ increases, the token retention rate rises, leading to improved Rouge-L and METEOR scores. However, these performance gains gradually plateau, indicating that the primary tokens identified by our nucleus selection strategy have already captured the most critical diagnostic features. This confirms that further increasing the token count yields diminishing returns, validating the efficiency of our adaptive selection mechanism.

\section{Conclusion}
In this paper, we propose MedPruner, a training-free hierarchical pruning framework that bridges the gap between high-performance 3D medical VLMs and the constraints of real-time clinical deployment. By adaptively managing the non-uniform information density across volumetric data, our approach ensures that critical diagnostic details are prioritized over redundant anatomical backgrounds. Extensive evaluations show that MedPruner improves the efficiency-performance trade-off without compromising diagnostic integrity, offering a scalable and model-agnostic solution for the practical integration of VLMs into complex medical workflows.

\noindent \textbf{Acknowledgments.} This work was supported by National Natural Science Foundation of China (No. 62522124), Hong Kong Research Grants Council General Research Fund 14206725 and Hong Kong Innovation and Technology Commission Innovation and Technology Fund PRP/082/24FX. 

\noindent \textbf{Disclosure of Interests.} The authors have no competing interests to declare that
are relevant to the content of this article.
\bibliographystyle{splncs04}
\bibliography{reference}

\end{document}